\documentclass{article}

\usepackage{microtype}
\usepackage{graphicx}
\usepackage{subcaption}
\usepackage{booktabs} 
\usepackage{array}
\usepackage{multirow}
\usepackage{amssymb}
\usepackage{xcolor}
\usepackage{pifont}
\usepackage[table]{xcolor}

\newcommand{\cmark}{\textcolor{green!60!black}{\ding{51}}}
\newcommand{\xmark}{\textcolor{red!70!black}{\ding{55}}}

\usepackage{hyperref}

\usepackage[accepted]{icml2026/icml2026}

\usepackage{amsmath}
\usepackage{amssymb}
\usepackage{mathtools}
\usepackage{amsthm}

\usepackage[capitalize,noabbrev]{cleveref}

\theoremstyle{plain}

\theoremstyle{definition}

\theoremstyle{remark}

\usepackage{xcolor}

\usepackage[textsize=tiny]{todonotes}

\icmltitlerunning{}

\begin{document}

\twocolumn[
  \icmltitle{DECAF: De-Clustering for Adaptive Representational Unlearning}



  \icmlsetsymbol{equal}{*}

  \begin{icmlauthorlist}
    \icmlauthor{Anjie Le}{ox}
    \icmlauthor{Can Peng}{ox}
    \icmlauthor{Hongcheng Guo}{fd}
    \icmlauthor{J. Alison Noble}{ox}

  \end{icmlauthorlist}

  \icmlaffiliation{ox}{Institute of Biomedical Engineering, University of Oxford, UK}
  \icmlaffiliation{fd}{Fudan University}

  \icmlcorrespondingauthor{Anjie Le}{anjie.le@eng.ox.ac.uk}
  \icmlcorrespondingauthor{J. Alison Noble}{alison.noble@eng.ox.ac.uk}

  \icmlkeywords{Machine Learning, ICML}

  \vskip 0.3in
]



\printAffiliationsAndNotice{}  
\vspace{-7mm}
\begin{abstract}
Machine unlearning, which aims to remove the influence of specific training data from a trained model, is a key requirement for privacy, accountability, and adaptive deployment. 
We argue that many unlearning methods are vulnerable to a simple \emph{clustering attack}, which can recover class structure in an unsupervised manner, limiting their suitability for continual deployment where removal requests must be handled reliably on demand.
To address this, we propose \textbf{DECAF} (DE-Clustering for Adaptive Forgetting), a post-hoc method that operates only on the forget set and is designed to break the cluster. 
DECAF combines input noise, confidence suppression, and entropy-based output diversification to disrupt the residual feature-space structure associated with forgotten data.
On CIFAR-10 with ResNet-18, DECAF attains 0.10\% forget-class accuracy, 79.4\% retain accuracy, and an AUS of 0.88 surpassing all other baselines.
In cluster-based analysis, it attains performance comparable to that of unlearning methods that use the full training set, while being significantly more efficient. 
Code: \url{https://github.com/ale256/representation_unlearning}.
\end{abstract}

\vspace{-7mm}

\section{Introduction}\label{sec:intro}

Foundation models are increasingly deployed in dynamic environments where their behavior must be updated after training. In such settings, adaptation is not limited to adding new capabilities: models must also {remove} outdated, harmful, or private information in a targeted and computationally sustainable manner. This requirement naturally arises in {adaptive deployment}, where data distributions and usage constraints evolve over time. As a result, machine unlearning becomes a core component of {continual adaptation}, rather than merely a post-deployment compliance tool.

A key instance of this problem is {deliberate forgetting}: given a trained model and a target {forget set}, the goal is to remove the influence of those examples while preserving utility. Retraining from scratch without the forget set \cite{cao2015towards} is the standard solution but is impractical at scale, motivating approximate unlearning methods that update a trained model while preserving performance on the remaining {retain set} \cite{bourtoule2021machine, lee2025esc}; however, many such methods assume access to the original data or pre-training corpora, which is often unrealistic \cite{gdpr, pipl2021, ccpa2018,bommasani2021opportunities, carlini2020extracting}.
This motivates the {post-hoc, {forget-only}} setting, where adaptation must be performed using only the data to be removed.


Beyond this practical limitation, we identify a more fundamental issue in existing unlearning methods. Unlearning is typically evaluated by reduced classification accuracy on the forget set, but this metric alone does not guarantee that the underlying information has been removed from the model \cite{golatkar_eternal_2020}. 
Consistent with prior evidence that class signal can live on in features \cite{influenceunlearn, ginart2019making}, we find that common unlearning methods still leave the forget set tightly clustered in the penultimate layer, indicating that class-discriminative information is 
preserved. This enables a simple {clustering attack}, in which an adversary can recover class structure from latent features with unsupervised clustering, revealing that many methods suppress outputs without erasing representations.

To address this, we propose \textbf{DECAF} (\textit{DE-Clustering for Adaptive Forgetting}), a lightweight post-hoc method that operates solely on the forget set. DECAF combines input noise, confidence suppression, and entropy-based output diversification, with each component acting on a complementary aspect of cluster geometry: they increase intra-class dispersion, weaken class-aligned representations, and encourage redistribution across the remaining classes, thereby explicitly disrupting residual feature-space structure and preventing re-clustering.

Our contributions are threefold. \emph{(i)}~We characterize a {clustering-based} failure mode: many post-hoc unlearning methods lower forget-class accuracy but leave the forget set class-structured in the penultimate layer, so a simple unsupervised clustering procedure can still recover class information, motivating evaluation beyond standard classification tests. \emph{(ii)}~We use clustering quality metrics to assess whether latent structure of the forget set has been disrupted, complementing standard unlearning measures. \emph{(iii)}~We propose \textbf{DECAF} as a lightweight forget-only alternative to retraining that directly targets declustering in feature space. On CIFAR-10 with ResNet-18, it yields a strong aggregate forgetting--utility trade-off and declustering competitive with retrain-from-scratch, using only the forget set and far less compute than retraining and typical retain-set fine-tuning.

\begin{figure*}[t]
\vspace{-2mm}
\centering
\includegraphics[width=0.8\linewidth, trim=0 16cm 0 5cm, clip]{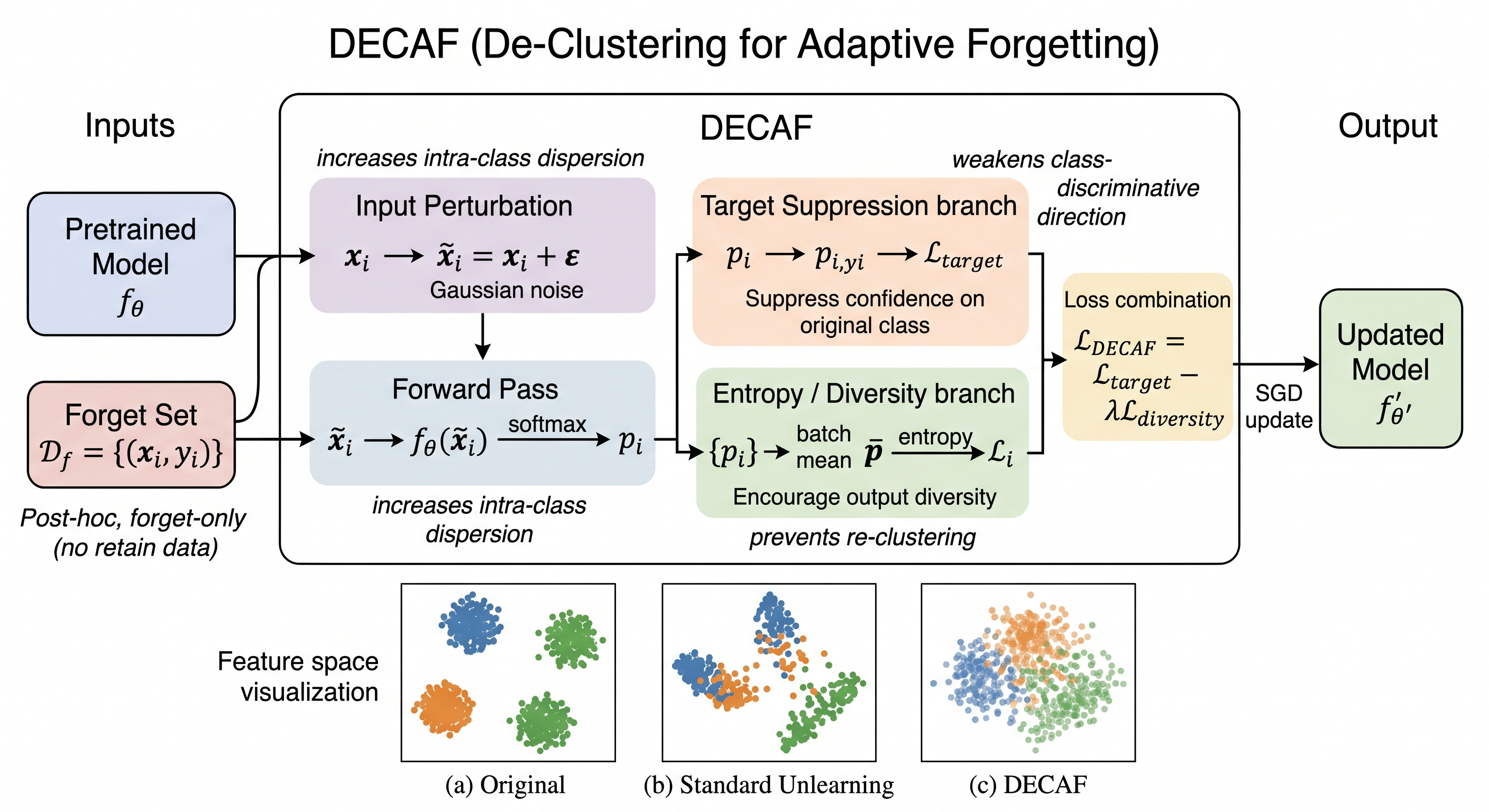}
\vspace{-1mm}
\caption{\textbf{DECAF overview.}
Given a pretrained model and a forget set $D_f$, DECAF performs post-hoc unlearning using only $D_f$. Inputs are perturbed with noise, passed through the model, and optimized with target suppression and entropy-based diversity, $\mathcal{L}_{\text{DECAF}} = \mathcal{L}_{\text{target}} - \lambda \mathcal{L}_{\text{diversity}}$.}
\vspace{-5mm}
\label{fig:main}
\end{figure*}

\section{Clustering Attack}

Most existing unlearning methods assess information removal by measuring classification degradation on the forget set $D_f$. 
However, poor classification performance on $D_f$ does not necessarily imply that its internal representation has been erased. 
In practice, we observe that even when the model misclassifies samples in $D_f$, their penultimate-layer features $z=\phi_{\theta'}(x)$ often remain tightly clustered and class-discriminative.
Prior work \cite{influenceunlearn, golatkar_eternal_2020, ginart2019making} has shown that such residual structure can often be recovered by training a linear probe on $Z_f$. While effective, linear probing requires data labels and an additional training stage, which is costly and sometimes impractical due to limited access to complete data. Our goal is instead to diagnose residual memorization directly from feature geometry, without additional fine-tuning.
\vspace{-1mm}

\subsection{Representation-Level Diagnosis}
Let $\mathcal{C}$ denote the set of classes and $Z_f^{(c)} = \{ \phi_{\theta'}(x) \mid x \in D_f, y(x) = c \}$ denote the latent features of the forget set after unlearning. 
We observe that for many baselines, these features remain compact within class and well separated across classes, i.e., for forget class $c$,
\vspace{-1mm}
\[
\mathbb{E}_{z,z' \in Z_f^{(c)}} \| z - z' \|^2 \ll 
\mathbb{E}_{z \in Z_f^{(c)}, z' \in Z_f^{(c')}} \| z - z' \|^2, \quad c' \ne c, 
\]
indicating that substantial class structure persists in feature space. In other words, unlearning may suppress the classifier output without fully removing the underlying geometry of the forgotten data.
\vspace{-1mm}

\subsection{Clustering-Based Evaluation}
 Accordingly, \textbf{we propose a clustering-based analysis} that directly examines the structure of latent representations as a diagnostic for residual memorization. Specifically, we cluster $Z_f$ under two distinct settings in an unsupervised manner, with number of classes $k=|\mathcal{C}|$ and $k=|\mathcal{C}|-1$ respectively, corresponding to whether the forgotten class remains a separable cluster or has been dispersed into the remaining classes. We apply multiple clustering algorithms (e.g., DBSCAN, K-Means, GMM) and choose the best performing one, then evaluate the resulting cluster assignments $\tilde{y}$ using three metrics:
\begin{itemize}
  \item \textbf{Silhouette Score:}
  \[
  \mathrm{Sil}(Z_f, \tilde{y}) = \frac{1}{n} \sum_{i=1}^n \frac{b(i) - a(i)}{\max\{a(i), b(i)\}},
  \]
  where $a(i)$ is the average distance between points in the same cluster, and $b(i)$ is the minimum average distance to points in any other cluster, measuring how well-separated and coherent each cluster is.

  \item \textbf{Calinski–Harabasz (CH) Index:}
  \[
  \mathrm{CH}(Z_f, \tilde{y}) = \frac{\mathrm{Tr}(B_k)}{\mathrm{Tr}(W_k)} \cdot \frac{n - k}{k - 1},
  \]
  where $\mathrm{Tr}(B_k)$ is the between-cluster dispersion, $\mathrm{Tr}(W_k)$ is the within-cluster dispersion, quantifing the ratio of between-cluster to within-cluster variance.

  \item \textbf{Davies–Bouldin (DB) Index:}
  \[
  \mathrm{DB}(Z_f, \tilde{y}) = \frac{1}{k} \sum_{i=1}^k \max_{j \ne i} \left( \frac{\sigma_i + \sigma_j}{d(c_i, c_j)} \right),
  \]
  where $\sigma_i$ is the average distance of points in cluster $i$ to its centroid $c_i$, and $d(c_i, c_j)$ is the distance between centroids, capturing intra-cluster tightness versus inter-cluster separation.

\end{itemize}
These metrics capture complementary aspects of clustering quality. 
Higher Silhouette and CH scores and lower DB scores indicate better-defined cluster structure.

\begin{table}[t]
    \centering
    \caption{Clustering quality of $D_f$ feature representations before and after unlearning. Lower Silhouette and CH scores and higher DB scores after unlearning indicate disrupted clustering and better forgetting. ``Safe?'' indicates whether a clustering attack can recover class identity.}
    \label{tab:clustering}
    \vspace{-2mm}
    \resizebox{0.48\textwidth}{!}{%
    \begingroup
    \setlength{\tabcolsep}{4.5pt}%
    \begin{tabular}{lcc>{\centering\arraybackslash}p{2em}cc>{\centering\arraybackslash}p{2em}cc>{\centering\arraybackslash}p{2em}c}
    \toprule
    \multirow{2}{*}{\textbf{Method}} &
    \multicolumn{3}{c}{\textbf{Silhouette} $\uparrow$} &
    \multicolumn{3}{c}{\textbf{Calinski--Harabasz} $\uparrow$} &
    \multicolumn{3}{c}{\textbf{Davies--Bouldin} $\downarrow$} &
    \multirow{2}{*}{\textbf{Safe?}} \\
    \cmidrule(lr){2-4} \cmidrule(lr){5-7} \cmidrule(lr){8-10}
    & $k=9$ & $k=10$ & $\Delta$ & $k=9$ & $k=10$ & $\Delta$ & $k=9$ & $k=10$ & $\Delta$ & \\
    \midrule
    \multicolumn{11}{c}{\textit{Methods requiring the retain set}} \\
    \midrule
    Retrain & 0.106 & 0.102 & \textcolor{green!40!black}{\scriptsize-0.004} & 1136.6 & 1048.5 & \textcolor{green!40!black}{\scriptsize-88.1} & 2.141 & 2.232 & \textcolor{green!40!black}{\scriptsize+0.091} & \cmark \\
    FT   & 0.178 & 0.179 & \textcolor{red!70!black}{\scriptsize+0.001} & 1326.5 & 1267.5 & \textcolor{green!40!black}{\scriptsize-59.0} & 1.688 & 1.726 & \textcolor{green!40!black}{\scriptsize+0.038} & \xmark \\
    FCS  & \textbf{0.218} & \textbf{0.205} & \textcolor{green!40!black}{\scriptsize-0.013} & \textbf{1460.2} & \textbf{1414.0} & \textcolor{green!40!black}{\scriptsize-46.2} & \textbf{1.549} & \textbf{1.610} & \textcolor{green!40!black}{\scriptsize+0.061} & \cmark \\
    MSG  & 0.085 & 0.080 & \textcolor{green!40!black}{\scriptsize-0.005} & 642.2  & 589.5  & \textcolor{green!40!black}{\scriptsize-52.7} & 2.579 & 2.568 & \textcolor{red!70!black}{\scriptsize-0.011} & \xmark \\
    \midrule
    \multicolumn{11}{c}{\textit{Methods using forget set only}} \\
    \midrule
    GA   & 0.052 & 0.049 & \textcolor{green!40!black}{\scriptsize-0.003} & 310.5  & 241.7  & \textcolor{green!40!black}{\scriptsize-68.8} & 3.021 & 2.997 & \textcolor{red!70!black}{\scriptsize-0.024} & \xmark \\
    \rowcolor{yellow!15}
    {DECAF} & \underline{0.185} & \underline{0.183} & \textcolor{green!40!black}{\scriptsize-0.002} & \underline{1340.8} & \underline{1283.5} & \textcolor{green!40!black}{\scriptsize-57.3} & \underline{1.682} & \underline{1.700} & \textcolor{green!40!black}{\scriptsize+0.018} & \cmark \\
    \bottomrule
    \end{tabular}%
    \endgroup
    }
    \vspace{-3mm}
  \end{table}

\subsection{Clustering Analysis}
\label{sec:experiments}

Residual clustering implies incomplete forgetting. As shown in Table~\ref{tab:clustering}, clustering-based analysis reveals several findings. Methods including FT, MSG and GA continue to exhibit recoverable structure, indicating that forgotten samples remain clustered in latent space despite reduced classification accuracy. This suggests that these approaches primarily suppress the output layer without fully removing underlying feature representations.
This analysis complements classification-based evaluation by revealing whether unlearning has truly disrupted the latent structure of $D_f$, rather than merely weakening the final decision boundary, which motivates our method that focuses on deeper representational forgetting.

\vspace{-1mm}

\section{Method}

{DECAF} is a post-hoc, forget-only unlearning method that operates using only the forget set $D_f$. 
It is designed to disrupt the three complementary properties of cluster structure. 
Input perturbation reduces cluster compactness, target suppression weakens separation from the original class boundary, and entropy regularization prevents forgotten samples from collapsing into a new surrogate cluster (Fig. \ref{fig:main}).

Given a forget example $(x_i, y_i) \in D_f$, we first perturb the input with Gaussian noise:
\begin{equation}
\tilde{x}_i = x_i + \epsilon, \qquad \epsilon \sim \mathcal{N}(0, \sigma^2 I),
\label{eq:gaussian}
\end{equation}
which disrupts low-level cues associated with memorized representations. From a clustering perspective, this increases the within-class dispersion of forgotten features and therefore weakens the compactness of the forget-class cluster.

Next, we suppress the model’s confidence in the forget-class label via
\begin{equation}
\mathcal{L}_{\text{target}} = \frac{1}{N} \sum_{i=1}^{N} p_{i,y_i},
\end{equation}
where $p_{i,y_i}$ is the softmax probability assigned to the ground-truth forget class $y_i$. 
Minimizing this objective discourages the model from preserving the original class-discriminative direction of forgotten samples, thereby destabilizing their shared representation and reducing their compactness and separability in feature space.

However, confidence suppression alone can lead to a degenerate solution in which forgotten samples are reassigned to a small subset of retained classes, forming a new cluster rather than being dispersed. 
To prevent this, we encourage diversity in the batch-average output distribution. Let
\[
\bar{p} = \frac{1}{N} \sum_{i=1}^{N} \mathrm{softmax}(f_\theta(\tilde{x}_i))
\]
denote the mean prediction over a minibatch. We define the diversity objective as
\begin{equation}
\mathcal{L}_{\text{diversity}} = - \sum_{c=1}^{C} \bar{p}_c \log \bar{p}_c,
\end{equation}
which promotes a more diffuse allocation of predictions across the remaining classes. In terms of cluster geometry, this discourages re-clustering and increases the overlap of forgotten samples with the rest of the feature space.

The final DECAF objective combines these two terms:
\begin{equation}
\mathcal{L}_{\text{DECAF}} = \mathcal{L}_{\text{target}} - \lambda \mathcal{L}_{\text{diversity}},
\label{eq:effect}
\end{equation}
where $\lambda$ controls the strength of output diversification. We optimize this objective using SGD for several epochs on $D_f$ only. This makes DECAF a lightweight and practical approach to forget-only representation-level unlearning.

\section{Experiment}

\subsection{Experimental Setup}

\paragraph{Setup.}
We evaluate DECAF on CIFAR-10 and randomly select one class as the forget set $D_f$, with the remaining samples forming the retain set $D_r$.
Model adapts a ResNet-18 backbone. Unlearning methods are applied post-hoc to the trained model.

\paragraph{Baselines.}
We compare DECAF with several post-hoc unlearning methods: Retrain-from-Scratch (Gold), which retrains using only $D_r$; Gradient Ascent (GA), which maximizes loss on $D_f$; Fine-Tuning (FT), which continue training the model for a few more epochs with $D_r$; Masked Small Gradient (MSG) \cite{cadet2024deep}, which suppresses updates on salient features; and Forget–Contrast–Strengthen (FCS)  \cite{cadet2024deep}, a method based on contrastive learning of representations.

\begin{table}[t]
    \centering
    \caption{Unlearning performance on CIFAR-10 with ResNet-18. 
    Lower Forget Acc, MIA, and AIN indicate stronger forgetting, while higher Retain Acc and AUS reflect better utility. 
    Bold entries indicate best results.}
    \vspace{-2mm}
    \label{tab:main_results}
    \resizebox{0.48\textwidth}{!}{
    \begin{tabular}{lcccccc}
    \toprule
    \textbf{Method} & \textbf{Forget Acc} $\downarrow$ & \textbf{Retain Acc} $\uparrow$ & \textbf{AUS} $\uparrow$ & \textbf{MIA} $\downarrow$ & \textbf{AIN} $\uparrow$ & \textbf{Time (s)} \\
    \midrule
    Retrain  & {0.00} & 77.29 & 0.86 & {51.70} & {1.00} & 1113.72 \\ 
    \midrule
    \multicolumn{7}{c}{\textit{Methods requiring the retain set}} \\
    \midrule
    FT  & 30.80 & \underline{82.48} & 0.72 & 59.70 & \underline{0.10} & 873.81 \\
    FCS & \underline{9.00} & \textbf{84.49} & \underline{0.77} & 69.30 & 0.04 & 139.78 \\
    MSG & 15.20 & 71.74 & 0.74 & \underline{55.80} & {0.02} & 106.09 \\
    \midrule
    \multicolumn{7}{c}{\textit{Methods using forget set only}} \\
    \midrule
    GA & 18.50 & 37.11 & 0.44 & \textbf{54.80} & \underline{0.10} & \textbf{2.42} \\
    \rowcolor{yellow!15}
    {DECAF} & \textbf{0.10} & {79.42} & \textbf{0.88} & 58.40 & \textbf{0.38} & \underline{9.55} \\
    \bottomrule
    \end{tabular}}
    \vspace{-4mm}
\end{table}

\paragraph{Metrics.}
We evaluate unlearning performance using standard metrics, including forget accuracy (lower is better), retain accuracy (higher is better), membership inference attack (MIA), and runtime. We also report an aggregated unlearning score (AUS) \cite{cotogni_duck_2024, li2025machine} to summarize the trade-off between forgetting and utility, and anamnesis index (AIN) \cite{chundawat2023zero} which measures residual memory of the forget data.

\subsection{Main Results}

Table~\ref{tab:main_results} presents the unlearning performance of DECAF alongside several baseline methods. 

\paragraph{Forgetting efficacy and utility trade-off.}
DECAF achieves the lowest forget accuracy (0.10\%) among all approximate methods, effectively eliminating class-specific predictive power. At the same time, it retains strong utility with a 79.4\% accuracy on the retain set, outperforming GA and MSG, and only slightly trailing FT and FCS. These two trends combine into an AUS of 0.88, higher than all baselines and even surpassing the retrain-from-scratch gold standard (0.86), demonstrating DECAF's superior ability to forget cleanly while preserving generalisation.

\paragraph{Latent memorisation and inference risk.}
In terms of MIA, DECAF balances between robustness and privacy, achieving stronger protection against MIA than FT and FCS, while preserving higher utility than GA. DECAF also obtains the highest AIN (0.38), indicating minimal residual influence from the forget set in internal representations. 

\paragraph{Efficiency.}
In terms of runtime, DECAF completes unlearning in under 10 seconds, which is over 100× faster than retraining and significantly more efficient than high-performing methods like FT and FCS. While GA is technically faster, its poor retain accuracy and low AUS indicate severe degradation of model utility, rendering it impractical for real-world use.

\definecolor{darkblue}{RGB}{31,119,180}
\begin{figure}[t]
    \centering
    \vspace{-1mm}
    \resizebox{0.45\textwidth}{!}{
    \begin{subfigure}[b]{0.32\linewidth}
        \includegraphics[width=\linewidth, trim=45 30 150 70, clip]
        {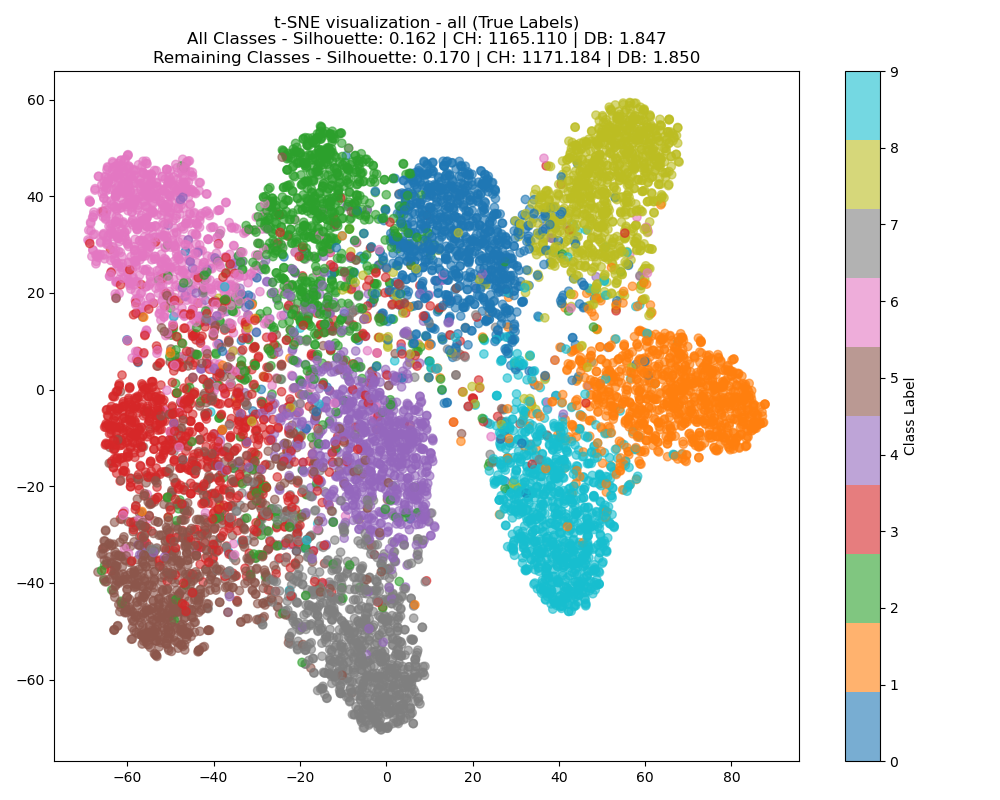}
        \caption{Original}
    \end{subfigure}
    \hfill
    \begin{subfigure}[b]{0.32\linewidth}
        \includegraphics[width=\linewidth, trim=45 30 150 70, clip]
        {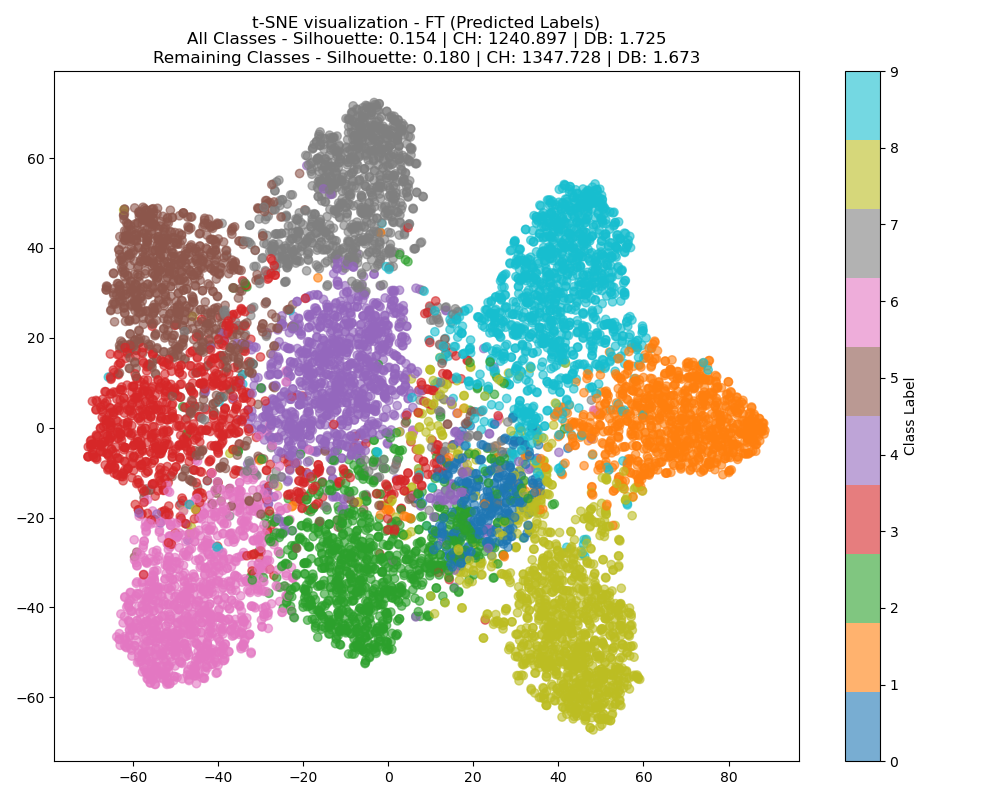}
        \caption{Fine-Tuning (FT)}
    \end{subfigure}
    \hfill
    \begin{subfigure}[b]{0.32\linewidth}
        \includegraphics[width=\linewidth, trim=45 30 150 70, clip]
        {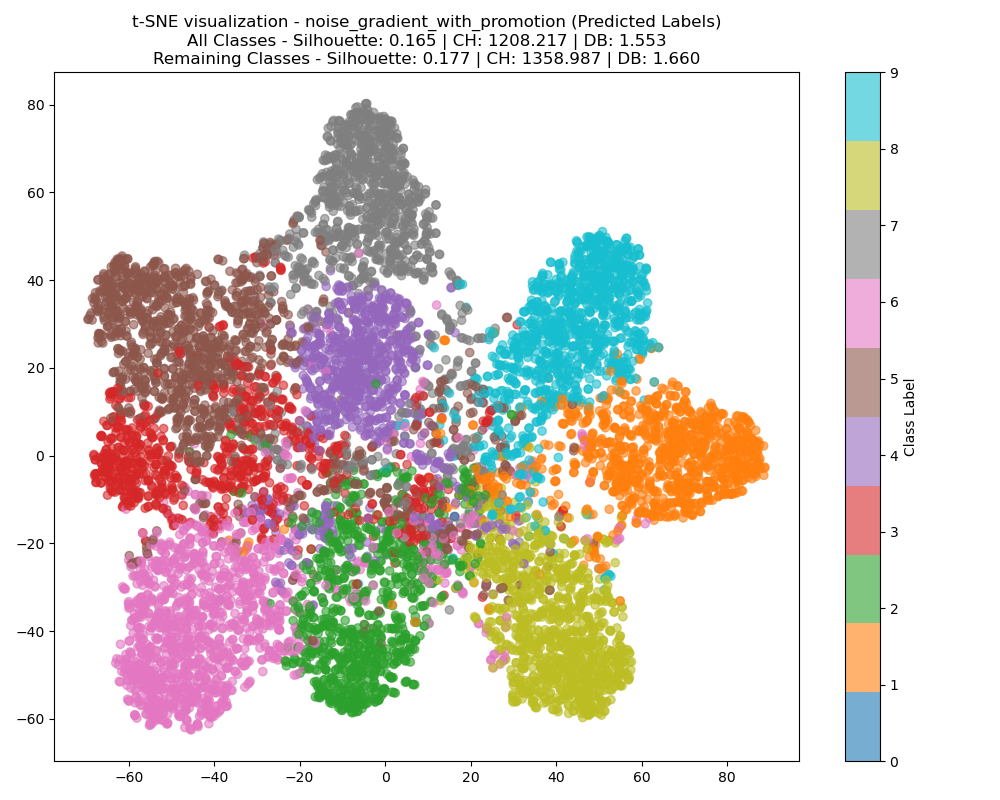}
        \caption{DECAF (Ours)}
    \end{subfigure}}
    \vspace{-1mm}
    \caption{t-SNE visualization of penultimate-layer features, where different colors indicate different predicted labels. The original model exhibits well-separated clusters. After fine-tuning, the forget class (\textcolor{darkblue}{dark blue}) remains clustered, indicating residual structure. In contrast, DECAF disperses the forget-class representations.}
    \vspace{-2mm}
    \label{fig:tsne}
\end{figure}

\paragraph{Declustering analysis.}
As shown in Table~\ref{tab:clustering}, {DECAF effectively disrupts cluster structure}. Compared to baselines, DECAF consistently reduces cluster separability across metrics, indicating that forgotten samples are no longer organized into a distinct group. A t-SNE visualisation is included in Fig.~\ref{fig:tsne} to illustrate this effect.

\subsection{Ablation Study}

\begin{table}[t!]
\vspace{-1mm}
\centering
\caption{Ablation study of DECAF components. Each component contributes meaningfully; removing any harms forgetting quality and increases vulnerability to attacks.}
\label{tab:ablation}
\vspace{-1mm}
\resizebox{0.35\textwidth}{!}{
\begin{tabular}{lcccc}
\toprule
\textbf{Variant} & \textbf{FAcc} $\downarrow$ & \textbf{RAcc} $\uparrow$ & \textbf{AUS} $\uparrow$ & \textbf{MIA} $\downarrow$ \\
\midrule
Full                & \textbf{0.10}  & \underline{79.42} & \underline{0.88} & \textbf{58.40} \\
w/o CS & 6.70 & 74.97 & 0.79 & \underline{60.90} \\
w/o Entropy         & \textbf{0.10} & \textbf{80.24} & \textbf{0.89} & 63.10 \\
w/o Noise           & 0.50 & 77.67 & 0.86 & 62.10 \\
\bottomrule
\end{tabular}}
\vspace{-5mm}
\end{table}

To understand the contribution of each component in DECAF, we perform an ablation study by removing individual mechanisms and measuring the impact on forgetting quality and robustness.
As shown in Table~\ref{tab:ablation}, each component of DECAF plays a complementary role. Removing confidence suppression leads to the largest drop in AUS, indicating its importance for utility. Removing entropy improves retain accuracy but increases MIA, suggesting weaker robustness. Similarly, removing noise slightly improves retention but worsens MIA, highlighting its role in disrupting feature structure. Overall, the full DECAF configuration best balances forgetting, utility, and robustness.

\section{Conclusion}

Our experiments show that DECAF effectively removes both predictive and representational traces of the forget set by directly disrupting feature-space structure, rather than merely suppressing outputs. This leads to more diffuse and less entangled representations, which may benefit downstream adaptation under distribution shift or continual learning. Future work could evaluate DECAF on broader benchmarks and further examine these effects.

\nocite{langley00}

\bibliography{main}
\bibliographystyle{icml2026/icml2026}

\newpage
\appendix
\onecolumn
\section{Related Work}
\subsection{Machine Unlearning}
 Machine Unlearning (MU)~\cite{sisa, cao2015towards, gdpr} aim to remove the influence of specific data from trained models to support privacy requirements such as the “right to be forgotten.” While retraining a model from scratch without the target data is a straightforward solution~\cite{amnesiac}, it is computationally expensive and often impractical.
Early MU efforts focused on traditional machine learning models, such as linear regression~\cite{baumhauer2022machine}, k-means~\cite{ginart2019making}, and SVMs~\cite{chen2019novel}, where convexity and simplicity allow exact or approximate unlearning. However, these approaches do not generalize well to complex visual data, and they are also incompatible with deep neural networks (DNNs), which lack the tractable properties exploited by traditional models~\cite{zhou2025decoupled}.
Recent DNN-based MU methods, as categorized by the survey in~\cite{xu2023machine}, are broadly divided into two categories: Data Reorganization and Model Manipulation.
\underline{Data Reorganization} methods~\cite{amnesiac, basaran2025certified, bourtoule2021machine} focus on modifying the training data to reduce the model’s dependence on specific samples. Some approaches~\cite{amnesiac, sun2025unlearning} perturb data labels or features to obscure the influence of the target samples; some approaches~\cite{bourtoule2021machine, chowdhury2024towards} partition the dataset into disjoint subsets to isolate the effect of forget requests—though such methods often impose restrictive assumptions on the training process; and others~\cite{basaran2025certified, cao2015towards} replace the original data with transformed surrogates to facilitate more efficient unlearning.
\underline{Model Manipulation} methods~\cite{cadet2024deep, jia2023model} directly modify model parameters after training. These techniques either adjust the weights to counteract the influence of the target samples~\cite{guo2020certified, schelter2019amnesia, cadet2024deep}, or prune parameters most correlated with the forget set~\cite{jia2023model}.
In this work, we focus on deep learning–based unlearning under practical constraints, where direct retraining or restrictive training assumptions are infeasible.
\subsection{Unlearning Feature Analysis}
Recently, Feature Manipulation has emerged as a promising direction in the unlearning literature. These methods aim to remove the influence of forgettable data by modifying the learned feature representations, rather than relying on retraining or directly altering model weights. They leverage knowledge distillation~\cite{zhou2025decoupled} or feature alignment~\cite{jia2024dimalign} to disentangle and suppress feature activations associated with the forget set while retaining general knowledge. For example, Zhou et al.~\cite{zhou2025decoupled} propose a decoupled distillation framework that decomposes the unlearning objective into forgetting and retention terms, allowing for masked knowledge transfer that preserves class-discriminative features unrelated to the forget class. Similarly, Jia et al.~\cite{jia2024dimalign} introduce dimensional alignment as a mechanism to align the latent manifolds of retained and forgotten samples, enabling unlearning via geometric regularization. Qin et al.~\cite{qin2024residual} further explore residual feature erasure in pre-trained models by applying low-rank adaptation (LoRA) to isolate and update only the task-specific residuals associated with the forget set. Le et al.~\cite{le2025pour} study selective removal of unwanted information while preserving useful representations, emphasizing the importance of fine-grained control in feature space. Other works target fine-grained attribute unlearning by suppressing sensitive features from the internal representations~\cite{guo2022efficient}, or by editing latent codes in generative models~\cite{moon2023feature}. These methods provide a generalizable and architecture-agnostic framework for unlearning, particularly suited to class-centric and representation-focused tasks. 

However, despite their emphasis on feature-space manipulation, most existing methods lack a rigorous, quantitative evaluation of how well the feature representations of the forget set have been erased. In contrast, we propose an unsupervised, clustering-based metric suite that explicitly measures the residual structure of forgotten features in the latent space—offering a principled and label-free diagnostic for representational forgetting.




\end{document}